\documentclass[10pt,twocolumn,letterpaper]{article}

\usepackage{times}
\usepackage{epsfig}
\usepackage{graphicx}
\usepackage{amsmath}
\usepackage{amssymb}
\usepackage{sidecap}
\usepackage{epstopdf}
\usepackage{multirow}
\usepackage{enumitem}

\usepackage{caption}
\usepackage{subcaption}

\newsavebox\CBox
\def\textBF#1{\sbox\CBox{#1}\resizebox{\wd\CBox}{\ht\CBox}{\textbf{#1}}}

\usepackage[pagebackref=true,breaklinks=true,letterpaper=true,colorlinks,bookmarks=false]{hyperref}
\usepackage[top = 2.00cm, bottom = 2.00cm, left = 1.95cm, right = 1.95cm]{geometry}

\begin{document}

\title{\fontsize{22}{30}\selectfont PIRM Challenge on Perceptual Image Enhancement \\ on Smartphones: Report \vspace{5mm}}

\author{
Andrey Ignatov, Radu Timofte,
Thang Van Vu, Tung Minh Luu, Trung X Pham, Cao Van Nguyen, \\
Yongwoo Kim, Jae-Seok Choi, Munchurl Kim, Jie Huang, Jiewen Ran, Chen Xing, Xingguang Zhou, \\
Pengfei Zhu, Mingrui Geng, Yawei Li, Eirikur Agustsson, Shuhang Gu, Luc Van Gool, Etienne de Stoutz, \\
Nikolay Kobyshev, Kehui Nie, Yan Zhao, Gen Li, Tong Tong, Qinquan Gao, Liu Hanwen, Pablo Navarrete \\
Michelini, Zhu Dan, Hu Fengshuo, Zheng Hui, Xiumei Wang, Lirui Deng, Rang Meng, Jinghui Qin, Yukai \\
Shi, Wushao Wen, Liang Lin, Ruicheng Feng, Shixiang Wu, Chao Dong, Yu Qiao, Subeesh Vasu, Nimisha \\
Thekke Madam, Praveen Kandula, A. N. Rajagopalan, Jie Liu, Cheolkon Jung
\thanks{A. Ignatov and R. Timofte (\{andrey,radu.timofte\}@vision.ee.ethz.ch, ETH Zurich) are the challenge organizers, while the other authors participated in the challenge. The Appendix contains the authors' teams and affiliations. PIRM 2018 Challenge webpage: \url{http://ai-benchmark.org}}
\vspace{5mm}
}

\date{}

\maketitle

\begin{abstract}
This paper reviews the first challenge on efficient perceptual image enhancement with the focus on deploying deep learning models on smartphones. The challenge consisted of two tracks. In the first one, participants were solving the classical image super-resolution problem with a bicubic downscaling factor of 4. The second track was aimed at real-world photo enhancement, and the goal was to map low-quality photos from the iPhone 3GS device to the same photos captured with a DSLR camera. The target metric used in this challenge combined the runtime, PSNR scores and solutions' perceptual results measured in the user study. To ensure the efficiency of the submitted models, we additionally measured their runtime and memory requirements on Android smartphones. The proposed solutions significantly improved baseline results defining the state-of-the-art for image enhancement on smartphones.

\end{abstract}

\section{Introduction}

The majority of the current challenges related to AI and deep learning for image restoration and enhancement~\cite{timofte2017ntire,timofte2018ntire,Ancuti_2018_CVPR_Workshops,Arad_2018_CVPR_Workshops,blau2018pirm,shoeiby2018pirm} are primarily targeting only one goal~--- high quantitative results measured by mean square error (MSE), peak signal-to-noise ratio (PSNR), structural similarity index (SSIM), mean opinion score (MOS) and other similar metrics. As a result, the general recipe for achieving top results in these competitions is quite similar: more layers/filters, deeper architectures and longer training on dozens of GPUs. However, one question that might arise here is whether often marginal improvements in these scores are actually worth the tremendous computational complexity increase. Maybe it is possible to achieve very similar perceptual results by using much smaller and resource-efficient networks that can run on common portable hardware like smartphones or tablets. This question becomes of special interest due to the uprise of many machine learning and computer vision problems directly related to these devices, such as image classification~\cite{szegedy2016rethinking,howard2017mobilenets}, image enhancement~\cite{ignatov2017dslr,ignatov2017wespe}, image super-resolution~\cite{dong2016image,timofte2014a+}, object tracking~\cite{wu2015object,huang2017speed}, visual scene understanding~\cite{li2009towards,cordts2016cityscapes}, face detection and recognition~\cite{li2015convolutional,schroff2015facenet}, etc.

The PIRM 2018 challenge on perceptual image enhancement on smartphones is the first step towards benchmarking resource-efficient architectures for computer vision and deep learning problems targeted at high perceptual results and deployment on mobile devices~\cite{ignatov2018ai}. It considers two classical computer vision problems~--- image super-resolution and enhancement, and introduces target performance metrics that are taking into account both networks' runtime, their quantitative and qualitative visual results. In the next sections we describe the challenge and the corresponding datasets, present and discuss the results and describe the proposed methods.

\section{PIRM 2018 challenge}

The PIRM 2018 challenge on perceptual image enhancement on smartphones has the following phases:
\begin{enumerate}
    \item[i] \textit{development:} the participants get access to the data;
    \item[ii] \textit{validation:} the participants have the opportunity to validate their solutions on the server and compare the results on the validation leaderboard;
    \item[iii] \textit{test:} the participants submit their final results, models, and factsheets.
\end{enumerate}

The PIRM 2018 challenge on perceptual image enhancement on smartphones consists of two different tracks described below.

\subsection{Track A: Image super-resolution}

\begin{figure}[t!]
\centering
\resizebox{1.0\linewidth}{!}
{
\includegraphics[width=1.0\linewidth]{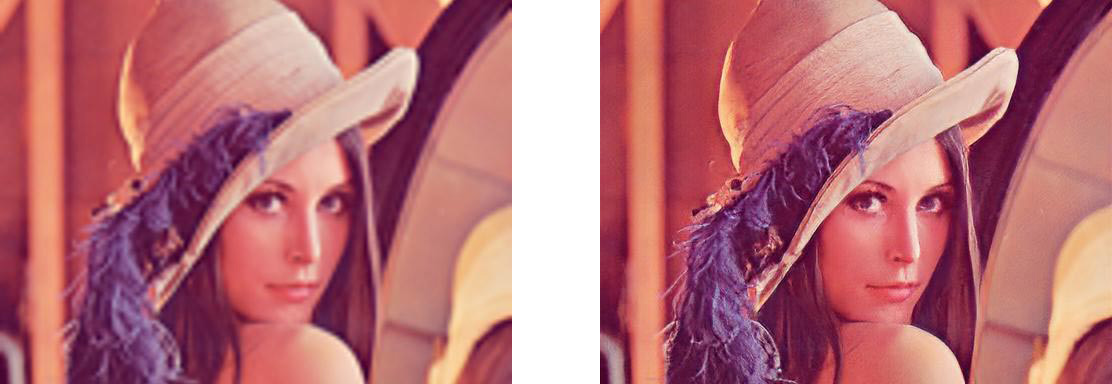}
}
\caption{\small{A low-res image (left) and the same image super-resolved by SRGAN (right).}}
\label{fig:superres}
\end{figure}

The first track is targeting a conventional super-resolution problem, where the goal is to reconstruct the original image based on its bicubically downscaled version. To make the task more practical, we consider a downscaling factor of 4, some sample results for which obtained with SRGAN network~\cite{ledig2017photo} are shown in the figure~\ref{fig:superres}. To train deep learning models, the participants used DIV2K dataset~\cite{agustsson2017ntire} with 800 diverse high-resolution train images crawled from the Internet.

\subsection{Track B: Image enhancement}

\begin{figure}[t!]
\centering
\resizebox{1.0\linewidth}{!}
{
\includegraphics[width=1.0\linewidth]{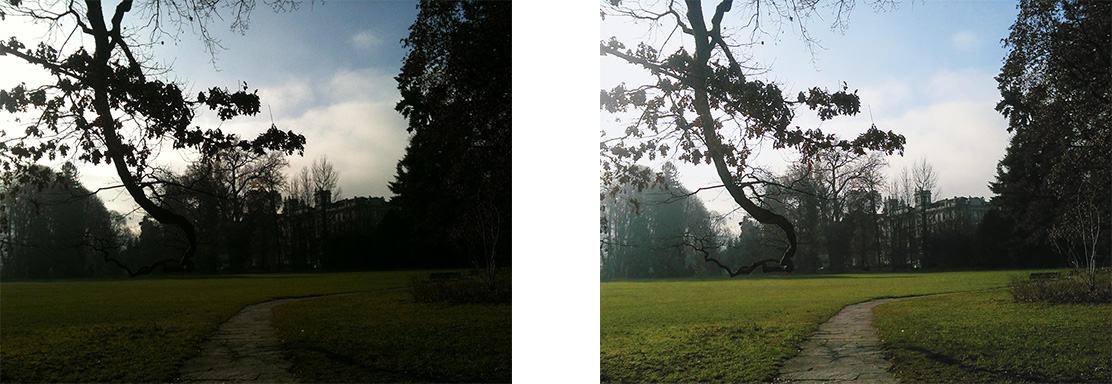}
}
\caption{\small{The original iPhone 3GS photo (left) and the same image enhanced by the DPED network~\cite{ignatov2017dslr} (right).}}
\label{fig:enhancement}
\end{figure}

The goal of the second track is to automatically improve the quality of photos captured with smartphones. In this task, we used DPED~\cite{ignatov2017dslr} dataset consisting of several thousands of images captured simultaneously with three smartphones and one high-end DSLR camera. Here we consider only a subtask of mapping photos from a very old iPhone 3GS device into the photos from Canon 70D DSLR. An example of the original and enhanced DPED test images are shown in the figure~\ref{fig:enhancement}.

\section{Scoring and validation}

The participants were required to submit their models as TensorFlow \textit{.pb} files that were later run on the test images and validated based on three metrics:

\begin{itemize}[label=$\bullet$]
\item Their speed on HD-resolution (1280$\times$720 pixels) images measured compared to the baseline SRCNN~\cite{dong2016image} network,

\smallskip

\item PSNR metric measuring their fidelity score,

\smallskip

\item MS-SSIM~\cite{wang2003multiscale} metric measuring their perceptual score.
\end{itemize}

Though MS-SSIM scores are known to correlate better with human image quality perception than PSNR, they are still often not reflecting many aspects of real image quality. Therefore, during the final test phase we conducted a user study involving more than 2000 participants (using MTurk platform~\footnote{\url{https://www.mturk.com/}}) that were asked to rate the visual results of all submitted solutions, and the resulting Mean Opinion Scores (MOS) then replaced MS-SSIM results. For Track B methods, the participants in the user study were invited to select one of four quality levels (probably worse, probably better, definitely better, excellent) for each method result in comparison with the original input image. The expressed preferences were averaged per each test image and then per each method to obtain the final MOS.

The final score of each submission was calculated as a weighted sum of the previous scores:

\footnotesize
\begin{equation*}
\begin{split}
\mathbf{Total \, Score} = \alpha \,\cdot\, (\mathrm{{PSNR_{solution}} \, - \, {PSNR_{\,\tiny{baseline}\,}}}) \, + \\
\, \beta \, \cdot \, (\mathrm{{MS\text{-}SSIM_{\,solution\,}} \, - \, {\, MS\text{-}SSIM_{\,\tiny{baseline}\,}}}) \, +  \\
\gamma \, \cdot \, \mathrm{min(4,\, {Time_{\,baseline\,}} \, / \, \, {Time_{\,solution\,}} )}.
\end{split}
\end{equation*}
\normalsize

\smallskip

To cover a broader range of possible targets, we have additionally introduced three validation tracks with different weight coefficients: the first one (score A) was favoring solutions with high quantitative results, the second one (score B)~--- with high perceptual results, and the third one (score C) was aimed at the best balance between the speed, visual and quantitative scores. Below are the exact coefficients for all tracks:

\smallskip

\textBF{Image super-resolution:}
\begin{itemize}[label=$\bullet$]
\item $\mathrm{PSNR_{\,\tiny{baseline}\,}}$ = 26.5, $\mathrm{SSIM_{\,\tiny{baseline}\,}}$ = 0.94,
\item $(\alpha, \beta, \gamma)$: \small{\, score A - $(4, 100, 1)$, \, score B - $(1, 400, 1)$,

\hspace{15.3mm} score C - $(2, 200, 1.5)$}.
\end{itemize}

\textBF{Image enhancement:}
\begin{itemize}[label=$\bullet$]
\item $\mathrm{PSNR_{\,\tiny{baseline}\,}}$ = 21.0, $\mathrm{SSIM_{\,\tiny{baseline}\,}}$ = 0.90,
\item $(\alpha, \beta, \gamma)$: \small{\, score A - $(4, 100, 2)$, \, score B - $(1, 400, 2)$,

\hspace{15.3mm} score C - $(2, 200, 2.9)$}.
\end{itemize}

The implementation of the scoring scripts, pre-trained baseline models and submission requirements are also available in the challenge github repository~\footnote{\url{https://github.com/aiff22/ai-challenge}}.

\section{Results}

\begin{table*}[t!]
\centering
\resizebox{2.0\columnwidth}{!}
{
\begin{tabular}{l|cc|cccc|c|ccc}
Team \, & \, PSNR \, & \, MS-SSIM \, & \, CPU, \, & \, GPU, \, & \, Razer Phone, \, & \, Huawei P20, \, & \, RAM \, & \, Score A \, & \, Score B \, & \, Score C \\
& & & \, ms \, & \, ms \, & \, ms \, & \, ms \, & & & & \\
\hline
\textBF{TEAM\_ALEX} \, & \, 28.21 \, & \, 0.9636 \, & \, 701 \, & \, 48 \, & \, 936 \, & \, 1335 \, & \, 1.5GB \, & \, \textbf{13.21} \, & \, \textbf{15.15} \, & \, \textbf{14.14} \\
\textBF{KAIST-VICLAB} \, & \, 28.14 \, & \, 0.9630 \, & \, \textbf{343} \, & \, \textbf{34} \, & \, \textbf{812} \, & \, \textbf{985} \, & \, 1.5GB \, & \, 12.86 \, & \, 14.83 \, & \, 13.87 \\
CARN\_CVL \, & \, 28.19 \, & \, 0.9633 \, & \, 773 \, & \, 112 \, & \, 1101 \, & \, 1537 \, & \, 1.5GB \, & \, 13.08 \, & \, 15.02 \, & \, 14.04 \\
IV SR+ \, & \, 28.13 \, & \, 0.9636 \, & \, 767 \, & \, 70 \, & \, 1198 \, & \, 1776 \, & \, 1.6GB \, & \, 12.88 \, & \, 15.05 \, & \, 13.97 \\
Rainbow	& 28.13 & 0.9632 & 654 & 56 & 1414 & 1749 & 1.5GB & 12.84 & 14.92 & \, 13.91 \\
Mt.Phoenix & 28.14 & 0.9630 & 793 & 90 & 1492 & 1994 & 1.5GB & 12.86 & 14.83 & \, 13.87 \\
SuperSR	& 28.18 & 0.9629 & 969 & 98 & 1731 & 2408 & 1.5GB & 12.35 & 14.17 & \, 12.94 \\
BOE-SBG	& 27.79 & 0.9602 & 1231 & 88 & 1773 & 2420 & 1.5GB & 9.79 & 11.98 & \, 10.55 \\
SRCNN (Baseline) \, & 27.21 & 0.9552 & 3239 & 205 & 7801 & 11566 & 2.6GB & 5.33 & 7.77 & \, 5.93 \\
\hline

\end{tabular}
}
\vspace{2.6mm}
\caption{\small{Track A (Image super-resolution), final challenge results.}}
\label{track-A}
\vspace{-1.2mm}
\end{table*}

During the validation phase, we have obtained more than 100 submissions from more than 20 different teams. 12 teams entered in the final test phase and submitted their models, codes and factsheets; tables~\ref{fig:superres} and~\ref{fig:enhancement} summarize their results.

\smallskip

\subsection{Image Super-Resolution}
First of all, we would like to note that all submitted solutions demonstrated high efficiency: they were generally three to eight times faster than SRCNN, and at the same time were providing radically better visual and quantitative results. Another interesting aspect is that according to the results of the user study, its participants were not able to distinguish between the visual results produced by different solutions, and MOS scores in all cases except for the baseline SRCNN model were almost identical. The reason for this is that neither of the submitted models were trained with a strong adversarial loss component: they were mainly optimizing Euclidean, MS-SSIM and VGG-based losses. In this track, however, we still have two winners: the first one is the solution proposed by TEAM\_ALEX that achieved the best scores in all three validation tracks, while the second winning solution from KAIST-VICLAB has demonstrated the best runtime on all platforms, including two Android smartphones (Razer Phone and Huawei P20) on which it was able to process HD-resolution images under 1 second.

\begin{table*}[tbh!]
\centering
\resizebox{2.0\columnwidth}{!}
{
\begin{tabular}{l|cc|c|cccc|c|ccc}
Team \, & \, PSNR \, & \, MS-SSIM \, & \, MOS \, & \, CPU, \, & \, GPU, \, & \, Razer Phone, \, & \, Huawei P20, \, & \, RAM \, & \, Score A \, & \, Score B \, & \, Score C \\
& & & & \, ms \, & \, ms \, & \, ms \, & \, ms \, & & & & \\
\hline
\textBF{Mt.Phoenix} \, & \, 21.99 \, & \, 0.9125 \, & \, \textbf{2.6804} \, & \, \textbf{682} \, & \, \textbf{64} \, & \, 1472 \, & \, 2187 \, & \, 1.4GB \, & \, \textbf{14.72} \, & \, \textbf{20.06} \, & \, \textbf{19.11} \\
EdS & 21.65 & 0.9048 & 2.6523&3241&253&5153& \, \tiny{Out of memory} \, &2.3GB&7.18&12.94& \, 9.36 \\
BOE-SBG&21.99&0.9079&2.6283&1620&111&1802&2321&1.6GB&10.39&14.61& \, 12.62 \\
MENet&22.22&0.9086&2.6108&1461&138&2279&3459&1.8GB&11.62&14.77& \, 13.47 \\
Rainbow&21.85&0.9067&2.5583&828&111&-&-&1.6GB&13.19&16.31& \, 16.93 \\
KAIST-VICLAB \, &21.56&0.8948&2.5123&2153&181&3200&4701&2.3GB&6.84&9.84& \, 8.65 \\
SNPR&22.03&0.9042&2.4650&1448&81&1987&3061&1.6GB&9.86&10.43& \, 11.05 \\
DPED (Baseline) \, &21.38&0.9034&2.4411& \, 20462 \, &1517&37003& \, \tiny{Out of memory} \, &3.7GB&2.89&4.90& \, 3.32 \\
Geometry&21.79&0.9068&2.4324&833&83&1209&1843&1.6GB&12.0&12.59& \, 14.95 \\
IV SR+&21.60&0.8957&2.4309&1375&125&1812&2508&1.6GB&8.13&9.26& \, 10.05 \\
SRCNN (Baseline) \, &21.31&0.8929&2.2950&3274&204&6890&11593&2.6GB&3.22&2.29& \, 3.49 \\
TEAM\_ALEX&21.87&0.9036&2.1196&781&70&962&1436&1.6GB&10.21&3.82& \, 10.81 \\
\end{tabular}
}
\vspace{2.6mm}
\caption{\small{Track B (Image enhancement), final results. The results are sorted according to the MOS scores. CNN model from Rainbow team was using \textit{tf.image.adjust\_contrast} operation not yet available in TensorFlow Mobile and was not able to run on Android.}}
\label{track-B}
\vspace{-1.2mm}
\end{table*}

\subsection{Image enhancement.}
Similarly to the previous task, all submissions here were able to significantly improve the runtime and PSNR scores of the baseline \mbox{SRCNN}~\cite{dong2016image,ignatov2017dslr} and DPED~\cite{ignatov2017dslr} approaches. Regarding the perceptual quality, in this case there is no clear story, mainly high PSNR scores did not guarantee the best visual results, and vice versa. Also, MS-SSIM does not predict well the perceptual quality captured by MOS. The winner of this track is Mt.Phoenix team that achieved top MOS scores, as well as the best A, B and C scores and the fastest runtime on CPU and GPU. On smartphones, this solution required around 1.5 and 2 seconds for enhancing one HD-resolution photo on the Razer Phone and Huawei P20, respectively.

\subsection{Discussion}
The PIRM 2018 challenge on perceptual image enhancement on smartphones promotes the efficiency in terms of runtime and memory as a critical measure for successful deployment of solutions on real applications and mobile devices.
For both considered tasks (super resolution and enhancement) a diversity of proposed solutions surpassed the provided baseline methods and greatly improved the efficiency.
We conclude that the challenge through the proposed solutions define the state-of-the-art for image enhancement on smartphones.

\section{Proposed methods}

\subsection{TEAM\_ALEX}
\label{ssc:TEAMALEX}

\begin{figure}[htb!]
\centering
\resizebox{0.94\linewidth}{!}
{
\includegraphics[width=\linewidth]{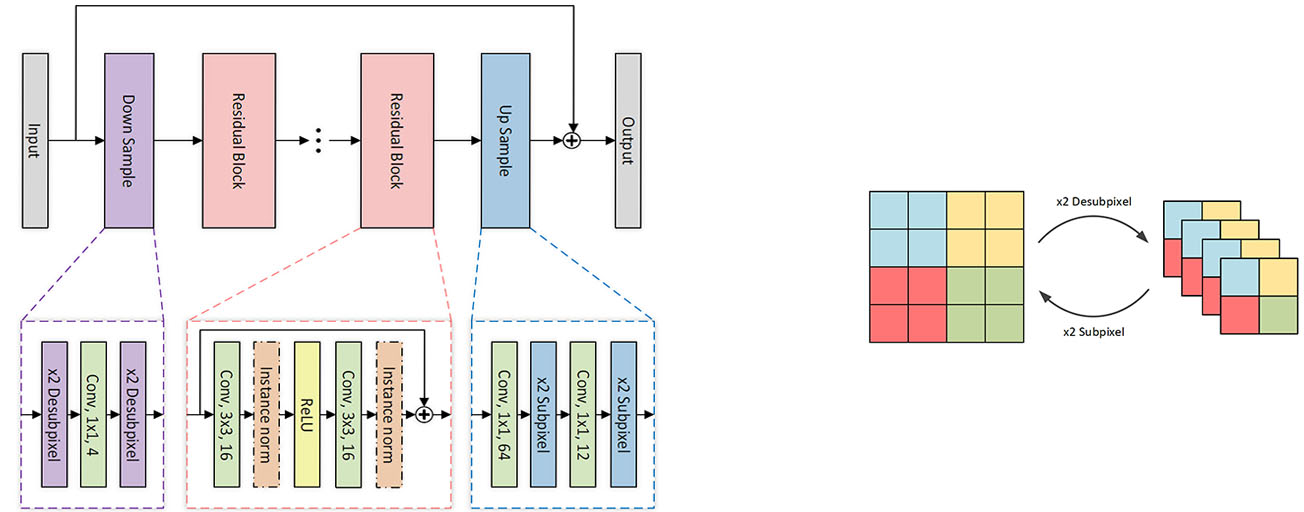}
}
\caption{\small{Desubpixel block and the CNN architecture proposed by TEAM\_ALEX.}}
\label{fig:TEAMALEX}
\end{figure}

This section describes solutions submitted by all teams participating in the final stage of the PIRM 2018 challenge on perceptual image enhancement on smartphones.

For track A, TEAM\_ALEX proposed a residual neural network with 20 residual blocks~\cite{vu2018fast}, though all computations in this CNN were mainly done on the images downscaled by a factor of 4 with two desubpixel blocks; in the last two layers they were upscaled back to their original resolution with two subpixel modules. The main idea of desubpixel downsampling is shown on the figure~\ref{fig:TEAMALEX}~--- this is a reversible downsampling done via rearranging the spatial features into several channels to reduce spatial dimensions without losing information. The whole network was trained with a combination of MSE and VGG-based loses on patches of size 196$\times$196px (image super-resolution) and 100$\times$100px (image enhancement) for 2$\times$10$^5$ and 2$\times$10$^6$ iterations, respectively. The authors used Adam optimizer with $\beta_1$ set to 0.9 and a batch size of 8; training data was additionally augmented with random flips and rotations. The learning rate was initialized at $1e-4$ and halved when the network was 60 percent trained.

\subsection{KAIST-VICLAB}
\label{ssc:KAIST-VICLAB}

\begin{figure}[htb!]
\centering
\resizebox{0.94\linewidth}{!}
{
\includegraphics[width=\linewidth]{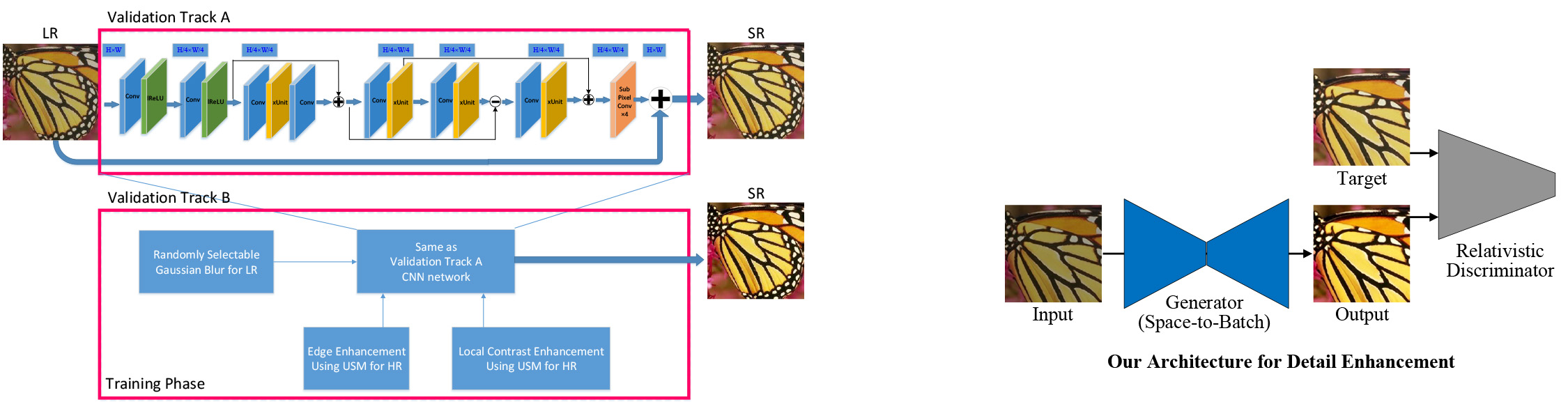}
}
\caption{\small{Solutions proposed by KAIST-VICLAB for tracks A (left) and B (right).}}
\label{fig:KAIST-VICLAB}
\end{figure}

In track A, KAIST-VICLAB proposed a similar approach of using 4$\times$ image downscaling and residual learning, however their CNN (fig.~\ref{fig:KAIST-VICLAB}) consisted of only 8 convolutional layers. High visual and quantitative results were still obtained by using a slightly different training scheme: the authors applied a small amount of Gaussian blur to degrade the downscaled low-resolution training patches, while they improved construct and sharpness of the target high-resolution images. Furthermore, residual units, pixel shuffle~\cite{shi2016real}, error feedback scheme~\cite{haris2018deep} and xUnit~\cite{kligvasser2017xunit} were integrated into network for faster learning and higher performance. The authors used 2,800 additional images from the BSDS300, Flickr500 and Flickr2K datasets for training, and augmented data with random flips and rotations. The network was trained for 2000 epochs on  128$\times$128px patches with L1 loss only; the batch size was set to 4, the learning rate was $1e-4$.

For track B, KAIST-VICLAB presented an encode-decoder based architecture (fig.~\ref{fig:KAIST-VICLAB}), where spatial sizes are reduced  with a space-to-batch technique: instead of using stride-2 convolutions, the feature maps obtained after each layer are divided into 4 smaller feature maps that are then concatenated along the batch dimension. The authors used an additional adversarial component, and for the discriminator they proposed relativistic RGAN~\cite{jolicoeur2018relativistic} with twice as many parameters as in the generator. The network was trained similarly to track A, but with a combination of color and adversarial losses from~\cite{ignatov2017dslr}.

\subsection{Mt.Phoenix}
\label{ssc:Mt.Phoenix}

\begin{figure}[htb!]
\centering
\resizebox{0.94\linewidth}{!}
{
\includegraphics[width=\linewidth]{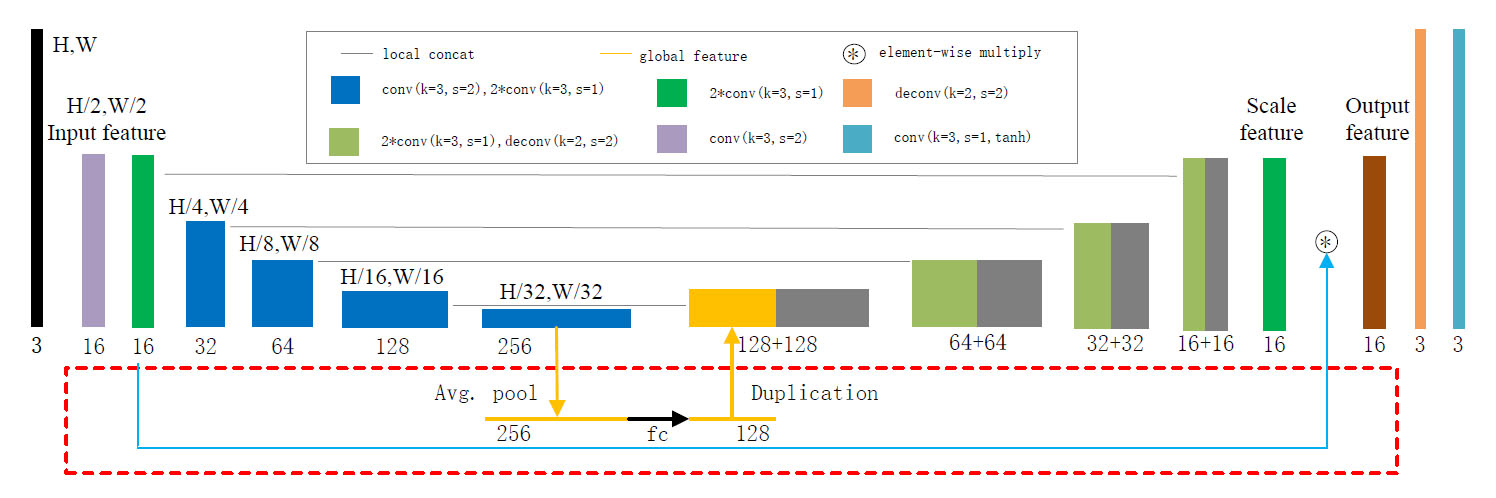}
}
\caption{\small{U-net architecture for image enhancement proposed by Mt.Phoenix.}}
\label{fig:Mt.Phoenix}
\end{figure}

For image super-resolution, the Mt.Phoenix authors used a deep residual CNN with two downsampling blocks performing image downscaling and two deconvolution blocks for its upscaling to the original size. Besides the standard residual blocks, additional skip connections between the input and middle layers were added to improve the performance of the network. CNN was trained on 500$\times$500px patches using Adam optimizer with an initial learning rate of $5e-4$ and a decay of $5e-5$. The network was trained with L1 loss, no data augmentation was used.

In the second track, Mt.Phoenix proposed a U-net style architecture~\cite{zhu2018range} (fig.~\ref{fig:Mt.Phoenix}) and augmented it with global features calculated by applying average pooling to features from its bottleneck layer. Additionally, a global transform layer performing element-wise multiplication of the outputs from the second and last convolutional layers was proposed. The network was trained with a combination of L1, MS-SSIM, VGG, total variation and GAN losses using Adam optimizer with a constant learning rate of $5e-4$.

\subsection{CARN\_CVL}
\label{ssc:CARNCVL}

\begin{figure}[htb!]
\centering
\resizebox{0.94\linewidth}{!}
{
\includegraphics[width=\linewidth]{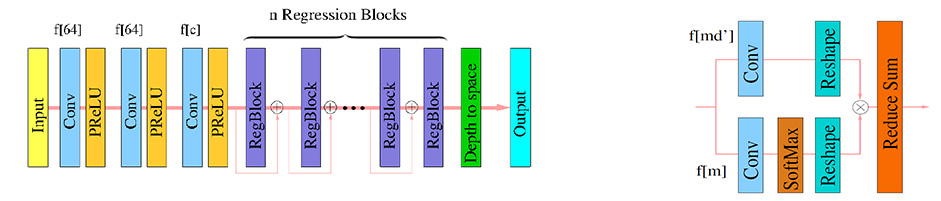}
}
\caption{\small{CARN architecture and CARN Regression Block presented by CARN\_CVL.}}
\label{fig:CARNCVL}
\end{figure}

For image super-resolution, CARN\_CVL proposed the convolutional anchored regression network (CARN)~\cite{li2018carn} (see Fig.~\ref{fig:CARNCVL}) which has the capability to efficiently trade-off between speed and accuracy. Inspired by A+~\cite{timofte2014a+,timofte2013anr} and ARN~\cite{agustsson2017arn}, CARN is formulated as a regression problem. The features are extracted from input raw images by convolutional layers. The regressors map features from low dimension to high dimension. Every regressor is uniquely associated with an anchor point so that by taking into account the similarity between the anchors and the extracted features, CARN can assemble the different regression results to form output features or the original image. In order to overcome the limitations of patch-based SR, all of the regressions and similarity comparisons between anchors and features are implemented by convolutional layers and encapsulated by a regression block. Furthermore, by stacking the regression block, the performance of the network increases steadily.
CARN\_CVL starts with the basic assumption of locally linear regression, derives the insights from it, and points out how to convert the architecture to convolutional layers in the proposed CARN.

The challenge entry uses CARN with 5 regression blocks, 16 anchors / regressors per block, and a number of feature layers reduced to 2. In the two feature layers, the stride of the convolution operation is set to 2 because the bicubic interpolated image contains no high frequency information compared to the LR image but slows down the executation of the network. The number of inner channels is set as 8 for the upscaling factor 4.

\subsection{EdS}
\label{ssc:EdS}

\begin{figure}[htb!]
\centering
\resizebox{0.94\linewidth}{!}
{
\includegraphics[width=\linewidth]{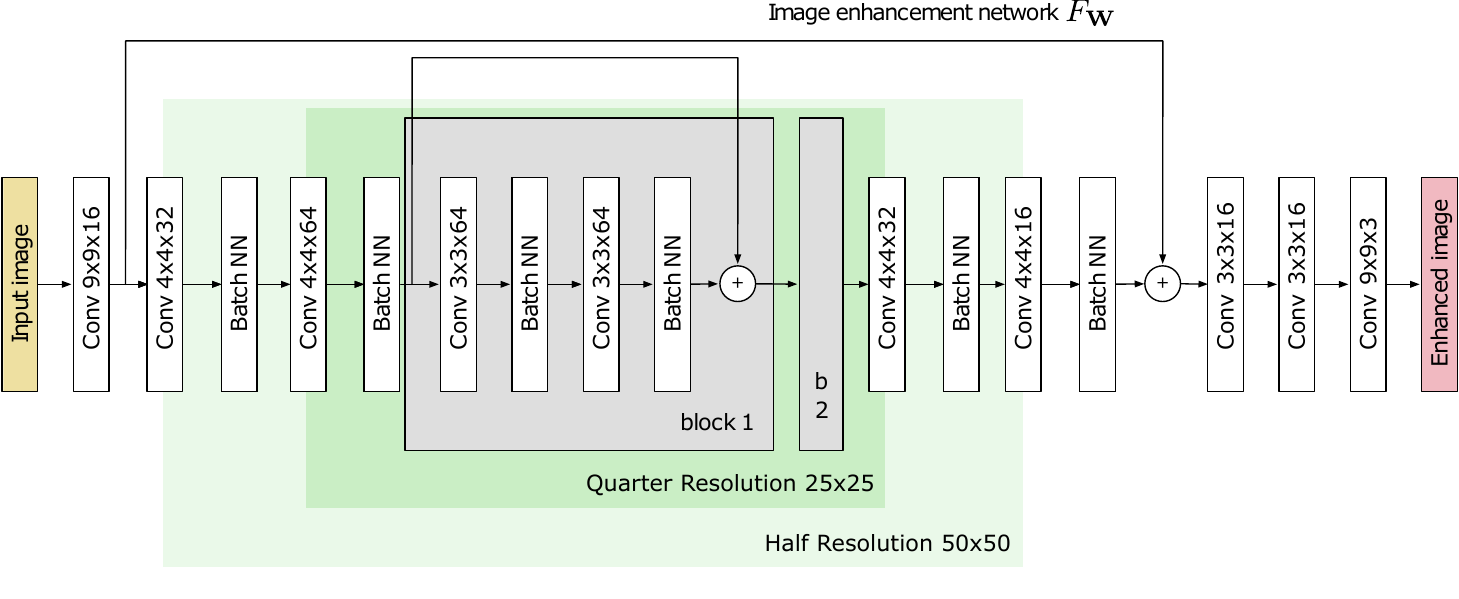}
}
\caption{\small{A variation of the original DPED architecture proposed by EdS team.}}
\label{fig:EdS}
\end{figure}

EdS proposed a modification~\cite{stoutz2018fast} of the original DPED ResNet architecture used for image enhancement (fig.~\ref{fig:EdS}). The main difference in their network was the use of two 4$\times$4 convolutional layers with stride 2 for going into lower dimensional space, and additional skip connections for faster training. The network was trained for 33K iterations using the same losses and setup as in~\cite{ignatov2017dslr}.

\subsection{IV SR+}
\label{ssc:IVSR}

\begin{figure}[htb!]
\centering
\resizebox{0.98\linewidth}{!}
{
\includegraphics[width=\linewidth]{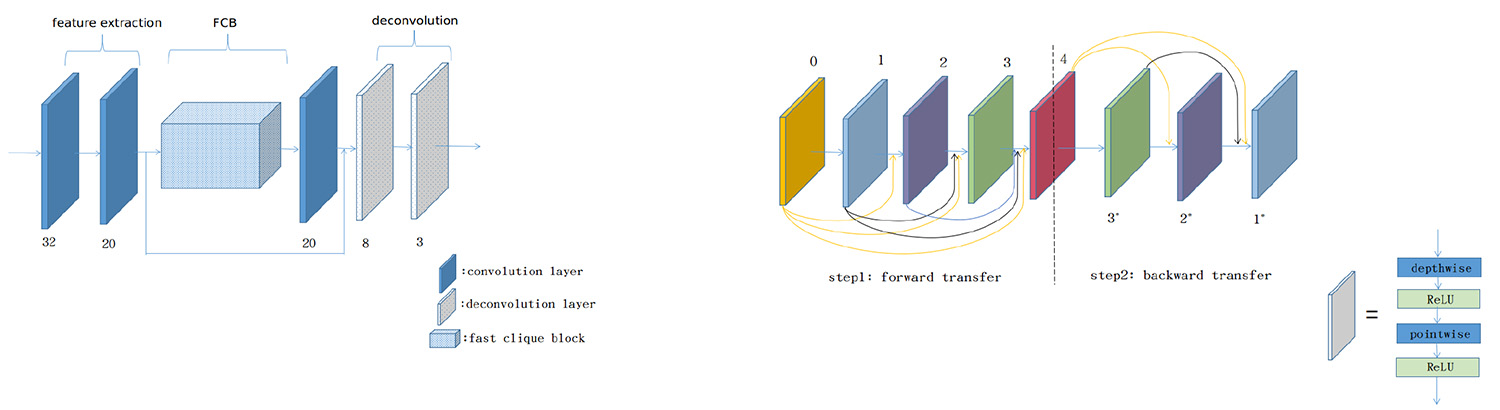}
}
\caption{\small{FCCN and the corresponding Fast Clique Block (FCB) proposed by IV SR+.}}
\label{fig:IVSR}
\end{figure}

The authors proposed a Fast Clique Convolutional Network (FCCN), which architecture was inspired by CliuqueNet~\cite{yang2018convolutional} and MobileNet~\cite{howard2017mobilenets}. The proposed FCCN consists of feature extraction, fast clique block (FCB) and two deconvolution layers (fig.~\ref{fig:IVSR}). For feature extraction, two convolutional layers with 32 and 20 kernels are utilized. Then, to accelerate the FCCN architecture, these features are fed to FCB layers for extracting more informative convolutional features. The FCB layer consists of one input convolutional layer and four bidirectional densely connected convolutional layers with both depthwise and pointwise convolution. The network was trained using Adam optimizer and a batch size of 16 for 3M iterations with an initial learning rate of $1e-4$ halved after 2 million iterations.

\subsection{BOE-SBG}
\label{ssc:BOE-SBG}

\begin{figure}[htb!]
\centering
\resizebox{0.94\linewidth}{!}
{
\includegraphics[width=\linewidth]{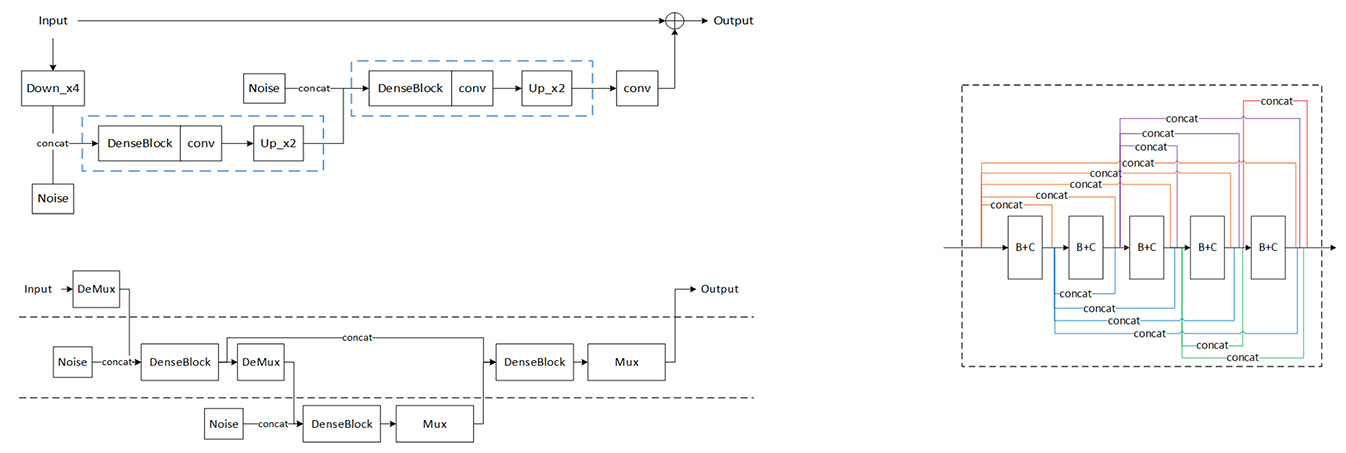}
}
\caption{\small{Neural networks for image super-resolution (top), image enhancement (bottom) and the corresponding Denseblock (right) proposed by BOE-SBG team.}}
\label{fig:BOE-SBG}
\end{figure}

The architecture of the network used for image super-resolution is presented in the figure~\ref{fig:BOE-SBG} and is based on the Laplacian pyramid framework with a denseblock inspired by~\cite{lai2017deep}. The parameters of denseblocks, strided and transposed convolutional layers are shared among different network levels to improve the performance. For image enhancement problem, the authors proposed a different architecture~\cite{liu2018deep} (fig.~\ref{fig:BOE-SBG}). First of all, it featured several Mux and Demux layers performing image up- and downscaling without information loss and that are basically a variant of (de)subpixel layers used in other approaches. This network was additionally trained with an extensive combination of various losses, including L1 loss for each image color channel, contextual, VGG, color, total variation and adversarial losses.

\subsection{Rainbow}
\label{ssc:Rainbow}

\begin{figure}[htb!]
\centering
\resizebox{0.94\linewidth}{!}
{
\includegraphics[width=\linewidth]{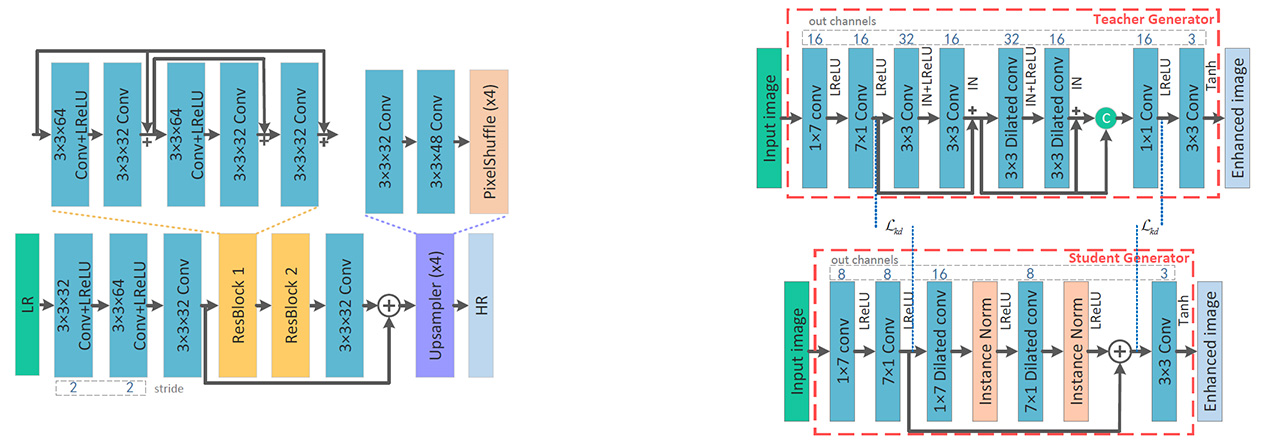}
}
\caption{\small{CNN architectures proposed by Rainbow for tracks A (left) and B (right).}}
\label{fig:Rainbow}
\end{figure}

The CNN architecture used in the first track is shown in the figure~\ref{fig:Rainbow}. The network consists of two convolutional layers with stride 2, three convolutional layers with stride 1, cascaded residual blocks and a subpixel layer. The network was trained to minimize L1 and SSIM losses on 384$\times$384px patches augmented with random flips and rotations. The learning rate was set to $5e-4$ and decreased by a factor of 5 every 1000 epochs.

A different approach~\cite{hui2018perception} was used for image enhancement: the authors first trained a larger teacher generator and then used it to guide the training of the smaller student network (see fig.~\ref{fig:Rainbow}). The latter was done by imposing additional knowledge distillation loss calculated as Euclidian distance between the corresponding normalized student's and teacher's feature maps. Besides this loss, the networks were trained with a combination of SSIM, VGG, L1, context, color and total variation losses using Adam optimizer with an initial learning rate of $5e-4$ decreased by a factor 10 for every $10^4$ iterations.

\subsection{MENet}
\label{ssc:MENet}

\begin{figure}[htb!]
\centering
\resizebox{0.97\linewidth}{!}
{
\includegraphics[width=\linewidth]{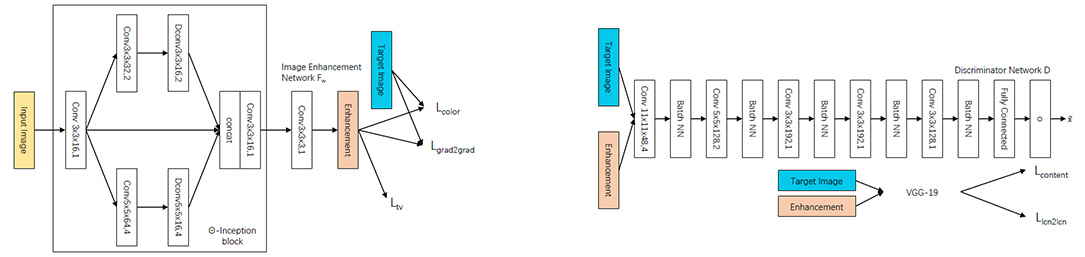}
}
\caption{\small{$\theta$-inception Network (generator and discriminator) presented by MENet team.}}
\label{fig:MENet}
\end{figure}

MENet team proposed a $\theta$-inception Network depicted in the figure~\ref{fig:MENet} for image enhancement problem. This CNN has a $\theta$-inception block where the image is processed in parallel by convolutional and deconvolutional layers with strides 2 and 4 for multi-scale learning. Besides that, the size of the convolutional filters is different too: 3 and 5 in the first and the second case, respectively. At the end of this block, the corresponding two outputs are concatenated together with the output from the first convolutional layer and are passed to the last CNN layer. The network is trained using the same setup as in~\cite{ignatov2017dslr} with the following two differences: 1) two additional texture loss functions (local contrast normalization and gradient) are used and 2) after pre-training the network is additionally fine-tuned on the same dataset with Adam minimizer and a learning rate of $1e-4$.

\subsection{SuperSR}
\label{ssc:SuperSR}

\begin{figure}[htb!]
\centering
\resizebox{0.94\linewidth}{!}
{
\includegraphics[width=\linewidth]{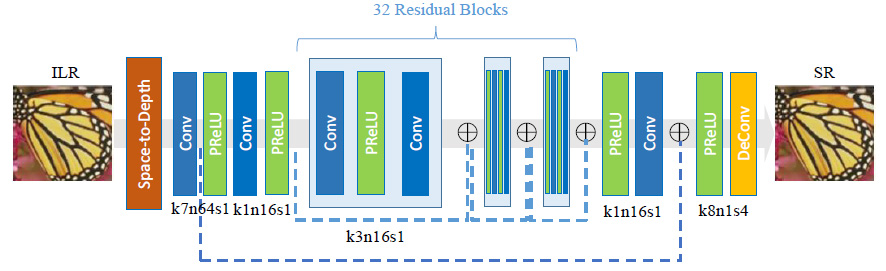}
}
\caption{\small{Deep residual network proposed by SuperSR team.}}
\label{fig:SuperSR}
\end{figure}

Figure~\ref{fig:SuperSR} presents the CNN architecture used for image super-resolution problem. The network consists of one space-to-depth 4$\times$ downsampling layer followed by convolutional and residual layers with PReLU activation functions and one deconvolutional layer for image upscaling. The model was trained on 192$\times$192px patches augmented with flips and rotations. Adam optimizer with a mini-batch size of 32 and a learning rate of $1e-3$ decayed by 10 every 1000 epochs was used for CNN training. After the initial pre-training with L2 loss, the training process was restarted with the same settings, while the loss function was replaced by a mixture of Charbonnier~\cite{barron2017more} loss and MS-SSIM losses.

\subsection{SNPR}
\label{ssc:SNPR}

\begin{figure}[htb!]
\centering
\resizebox{0.94\linewidth}{!}
{
\includegraphics[width=\linewidth]{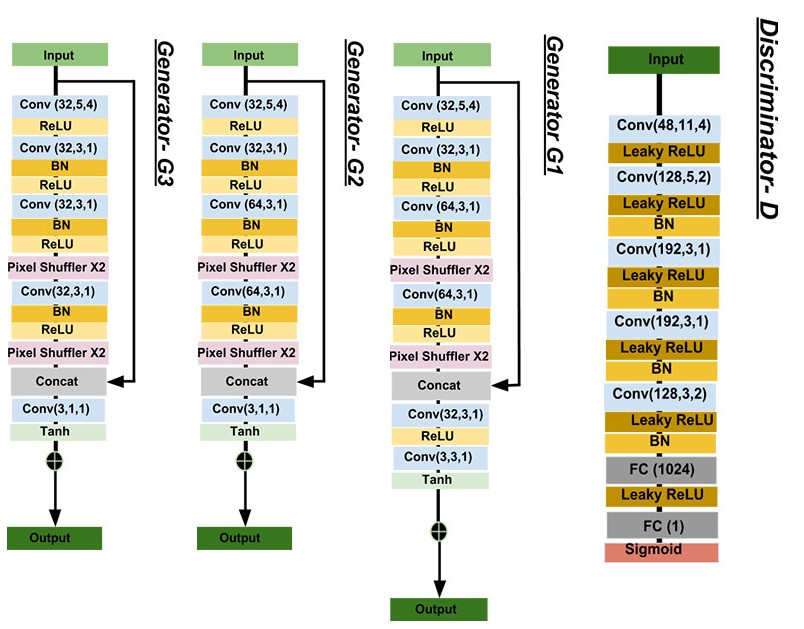}
}
\caption{\small{CNN architecture proposed by SNPR team.}}
\label{fig:SNPR}
\end{figure}

For image enhancement, SNPR derives three network architectures corresponding to different operating points.
The generator networks ($G1$, $G2$, and $G3$) corresponding to the three different approaches and the common discriminator network $D$ are shown in Fig.~\ref{fig:SNPR}. \textit{Conv(f, k, s)} refers to a convolution
layer with $f$  $k\times k$ filters performing convolution by a stride
factor of $s$, ReLU is a Rectified Linear Unit, BN refers to batch-normalization, and \textit{Pixel-Shuffler X2} refers to the pixel shuffler layer~\cite{shi2016real} which increases resolution by a factor of 2. The first three layers are
meant to extract the features that are relevant for image enhancement. Feature
extraction at low-image-dimension has the advantages of larger receptive field and much lower computational complexity~\cite{sim2018high}. To compensate for detrimental effects of spatial dimension reduction in features, the input image (which have full-resolution spatial features) is concatenated with the features extracted at low-dimensional space and then combined by the succeeding convolutional layers.
Overall $G3$ achieves the best speed-up-ratio but with a lower performance as compared to DPED baseline~\cite{ignatov2017dslr}, whereas $G1$ achieves the lowest speed-up-ratio while having comparable quality to that of DPED.

\subsection{Geometry}
\label{ssc:Geometry}

\begin{figure}[htb!]
\centering
\resizebox{0.84\linewidth}{!}
{
\includegraphics[width=\linewidth]{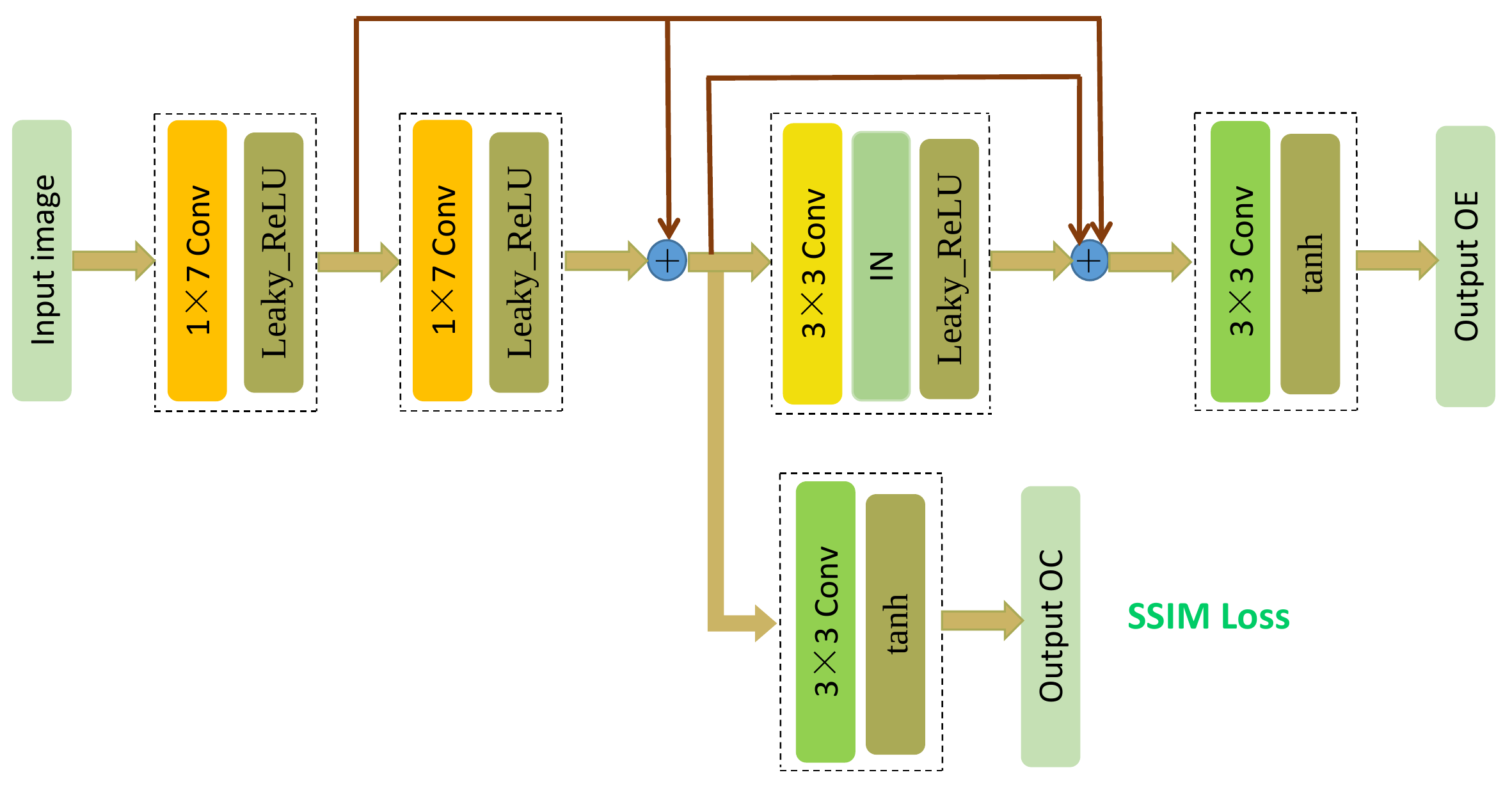}
}
\caption{\small{Neural network proposed by Geometry team.}}
\label{fig:Geometry}
\end{figure}

The overall structure of the network~\cite{liu2018multiple} presented by Geometry team is shown in the figure.~\ref{fig:Geometry}. Each convolutional layer has 16 filters, and the network itself produces two outputs: one based on the features from the middle CNN layer, and one from the last layer. The intermediate output (Output OC) is used to compute SSIM loss, while the final one (Output OE) is used to compute the loss function consisting of adversarial, smooth, and style losses. During the training all losses are summed, and the network is trained as a whole using Adam optimizer with a learning rate of $5e-4$ decreased by a factor of 10 every 8000 iterations.

\section*{Acknowledgements}

We  thank  the  PIRM2018  sponsors: ETH Zurich (Computer Vision Lab), Huawei Inc., MediaTek Inc., and Israel Institute of Technology.

\section*{Appendix 1: Teams and affiliations}
\label{sec:affiliations}

\smallskip


\textbf{PIRM 2018 Team}
\small
\bigskip

\noindent
\textBF{Title:} \hspace{0.7mm} PIRM Challenge on Perceptual Image Enhancement

\hspace{4.66mm} on Smartphones

\smallskip

\noindent
\textBF{Members:} \hspace{0.7mm} Andrey Ignatov \,--\, \footnotesize{andrey@vision.ee.ethz.ch}, \small

\hspace{10.9mm} Radu Timofte \,--\, \footnotesize{radu.timofte@vision.ee.ethz.ch}

\smallskip

\noindent
\textBF{Affiliations:} \hspace{0.7mm} Computer Vision Lab, ETH Zurich, Switzerland


\bigskip

\smallskip

\newpage

\noindent
\textbf{TEAM\_ALEX}
\small
\bigskip

\noindent
\textBF{Title:} \hspace{0.7mm} Fast and Efficient Image Quality Enhancement using

\hspace{4.66mm} Desubpixel Down-sampling~\cite{vu2018fast}

\smallskip

\noindent
\textBF{Members:} \hspace{0.7mm} Thang Vu \,--\, \footnotesize{thangvubk@kaist.ac.kr}, \small

\hspace{10.9mm} Tung Luu, Trung Pham, Cao Nguyen

\smallskip

\noindent
\textBF{Affiliations:} \hspace{0.7mm} Dept. of Electrical Engineering KAIST,

\hspace{13.2mm} Republic of Korea

\bigskip

\smallskip

\noindent
\textbf{KAIST-VICLAB}
\small
\bigskip

\noindent
\textBF{Title-A:} \hspace{0.7mm} A Low-Complexity Convolutional Neural Network for

\hspace{8.0mm} Perceptual Super-Resolution using Randomly-Selected

\hspace{8.0mm} Degraded LR and Enhanced HR

\smallskip

\noindent
\textBF{Title-B:} \hspace{0.7mm} A Convolutional Neural Network for Detail Enhancement

\hspace{8.0mm} with the Relativistic Discriminator

\smallskip

\noindent
\textBF{Members:} \hspace{0.7mm} Yongwoo Kim \,--\, \footnotesize{yongwoo.kim@kaist.ac.kr}, \small

\hspace{10.9mm} Jae-Seok Choi, Munchurl Kim

\smallskip

\noindent
\textBF{Affiliations:} \hspace{0.7mm} Video and Image Computing Lab, KAIST,

\hspace{13.2mm} Republic of Korea

\bigskip

\smallskip

\noindent
\textbf{Mt.Phoenix}
\small
\bigskip

\noindent
\textBF{Title-A:} \hspace{0.7mm} Multi Level Super Resolution Net~\cite{zhu2018range}

\smallskip

\noindent
\textBF{Title-B:} \hspace{0.7mm} Range Scaling Global U-Net for Perceptual Image

\hspace{8.0mm} Enhancement on Mobile Devices

\smallskip

\noindent
\textBF{Members:} \hspace{0.7mm} Pengfei Zhu \,--\, \footnotesize{zpf2@meitu.com}, \small

\hspace{10.9mm} Chen Xing, Xingguang Zhou, Jie Huang,

\hspace{10.9mm} Mingrui Geng, Jiewen Ran

\smallskip

\noindent
\textBF{Affiliations:} \hspace{0.7mm} Meitu Imaging \& Vision Lab, China

\bigskip

\smallskip

\noindent
\textbf{CARN\_CVL}
\small
\bigskip

\noindent
\textBF{Title:} \hspace{0.7mm} Convolutional Anchored Regression Network~\cite{li2018carn}

\smallskip

\noindent
\textBF{Members:} \hspace{0.7mm} Yawei Li \,--\, \footnotesize{yawei.li@vision.ee.ethz.ch}, \small

\hspace{10.9mm} Eirikur Agustsson, Shuhang Gu,

\hspace{10.9mm} Radu Timofte, Luc Van Gool

\smallskip

\noindent
\textBF{Affiliations:} \hspace{0.7mm} Computer Vision Lab, ETH Zurich, Switzerland

\bigskip

\smallskip

\noindent
\textbf{IV SR+}
\small
\bigskip

\noindent
\textBF{Title:} \hspace{0.7mm} An Efficient and Compact Mobile Image Super-resolution

\hspace{4.66mm} with Fast Clique Convolutional Network

\smallskip

\noindent
\textBF{Members:} \hspace{0.7mm} Kehui Nie $^1$ \,--\, \footnotesize{n161120080@fzu.edu.cn}, \small

\hspace{10.9mm} Yan Zhao $^1$, Gen Li $^2$, Tong Tong $^2$, Qinquan Gao $^1$

\smallskip

\noindent
\textBF{Affiliations:}  \hspace{0.7mm} $^1$~-- Fuzhou University, China

\hspace{13.2mm} $^2$~-- Imperial Vision, China

\bigskip

\noindent
\textbf{Rainbow}
\small
\bigskip

\noindent
\textBF{Title:} \hspace{0.7mm} Perception-Preserving Convolutional Networks for Image

\hspace{4.66mm} Enhancement on Smartphones~\cite{hui2018perception}

\smallskip

\noindent
\textBF{Members:} \hspace{0.7mm} Zheng Hui $^1$ \,--\, \footnotesize{zheng\_hui@aliyun.com}, \small

\hspace{10.5mm} Xiumei Wang $^1$, Lirui Deng $^2$, Rang Meng $^3$

\smallskip

\noindent
\textBF{Affiliations:}  \hspace{0.7mm} $^1$~-- Xidian University, China

\hspace{13.2mm} $^2$~-- Tsinghua University, China

\hspace{13.2mm} $^3$~-- Zhejiang University, China

\bigskip

\smallskip

\noindent
\textbf{SuperSR}
\small
\bigskip

\noindent
\textBF{Title:} \hspace{0.7mm} Enhanced FSRCNN for Image Super-Resolution

\smallskip

\noindent
\textBF{Members:} \hspace{0.7mm} Ruicheng Feng \,--\, \footnotesize{jnjaby@gmail.com}, \small

\hspace{10.9mm} Shixiang Wu, Chao Dong, Yu Qiao

\smallskip

\noindent
\textBF{Affiliations:} \hspace{0.7mm} Shenzhen Institute of Advanced Technology, China

\bigskip

\smallskip

\noindent
\textbf{BOE-SBG}
\small
\bigskip

\noindent
\textBF{Title-A:} \hspace{0.7mm} Deep Laplacian Pyramid Networks with Denseblock for

\hspace{8.0mm} Image Super-Resolution

\smallskip

\noindent
\textBF{Title-B:} \hspace{0.7mm} Deep Networks for Image-to-image Translation with

\hspace{8.0mm} Mux and Demux Layers~\cite{liu2018deep}

\smallskip

\noindent
\textBF{Members:} \hspace{0.7mm} Liu Hanwen \,--\, \footnotesize{liuhanwen@boe.com.cn}, \small

\hspace{10.9mm} Pablo Navarrete Michelini, Zhu Dan, Hu Fengshuo

\smallskip

\noindent
\textBF{Affiliations:} \hspace{0.7mm} BOE Technology Group Co., Ltd, China

\bigskip

\smallskip

\noindent
\textbf{EdS}
\small
\bigskip

\noindent
\textBF{Title:} \hspace{0.7mm} Fast Perceptual Image Enhancement~\cite{stoutz2018fast}

\smallskip

\noindent
\textBF{Members:} \hspace{0.7mm} Etienne de Stoutz \,--\, \footnotesize{etienned@ethz.ch}

\hspace{10.9mm} Nikolay Kobyshev

\smallskip

\noindent
\textBF{Affiliations:} \hspace{0.7mm} ETH Zurich, Switzerland

\bigskip

\smallskip

\noindent
\textbf{MENet}
\small
\bigskip

\noindent
\textBF{Title:} \hspace{0.7mm} Fast and Accurate DSLR-Quality Photo Enhancement

\hspace{4.66mm} Using $\theta$-inception Network

\smallskip

\noindent
\textBF{Members:} \hspace{0.7mm} Jinghui Qin \,--\, \footnotesize{qinjingh@mail2.sysu.edu.cn}, \small

\hspace{10.6mm} Yukai Shi, Wushao Wen, Liang Lin

\smallskip

\noindent
\textBF{Affiliations:} \hspace{0.7mm} Sun Yat-sen University, China

\bigskip

\smallskip

\noindent
\textbf{SNPR}
\small
\bigskip

\noindent
\textBF{Title:} \hspace{0.7mm} Efficient Perceptual Image Enhancement Network

\hspace{4.66mm}  for Smartphones

\smallskip

\noindent
\textBF{Members:} \hspace{0.7mm} Subeesh Vasu \,--\, \footnotesize{subeeshvasu@gmail.com}, \small

\hspace{10.9mm} Nimisha Thekke Madam, Praveen Kandula,

\hspace{10.9mm} A. N. Rajagopalan

\smallskip

\noindent
\textBF{Affiliations:} \hspace{0.7mm} Indian Institute of Technology Madras, India

\bigskip

\smallskip

\noindent
\textbf{Geometry}
\small
\bigskip

\noindent
\textBF{Title:} \hspace{0.7mm} Multiple Connected Residual Network for Image

\hspace{4.66mm} Enhancement on Smartphones~\cite{liu2018multiple}

\smallskip

\noindent
\textBF{Members:} \hspace{0.7mm} Jie Liu \,--\, \footnotesize{jieliu543@gmail.com}, \small

\hspace{10.7mm} Cheolkon Jung

\smallskip

\noindent
\textBF{Affiliations:} \hspace{0.7mm} School of Electronic Engineering,

\hspace{13.2mm} Xidian University, China

\bigskip

\smallskip

{\footnotesize
\bibliographystyle{splncs}

}

\end{document}